\documentclass[letterpaper]{article} 
\usepackage[preprint]{aaai2027}  
\usepackage[hyphens]{url}  
\usepackage{graphicx} 
\usepackage{natbib}  
\usepackage{caption} 
\usepackage{algorithm}
\usepackage{algorithmic}
\usepackage{amsmath}
\usepackage{amssymb}
\usepackage{graphicx}
\usepackage{multirow}
\usepackage{newfloat}
\usepackage{listings}
\DeclareCaptionStyle{ruled}{labelfont=normalfont,labelsep=colon,strut=off} 
\floatstyle{ruled}
\newfloat{listing}{tb}{lst}{}
\floatname{listing}{Listing}

\usepackage{booktabs}

\title{SPFM-Net: Semantic-Prior-Guided Frequency-Constrained Mamba for Invisible Watermark Attack }
\author{
    Chunpeng Wang\textsuperscript{\rm 1},
    Yanan Shi\textsuperscript{\rm 1},
    Zhiqiu Xia\textsuperscript{\rm 2},
    Jidong Yang\textsuperscript{\rm 1},
    Suo Gao\textsuperscript{\rm 3},
    Qi Li\textsuperscript{\rm 1}\corresponding
}
\affiliations{
    \textsuperscript{\rm 1}Qilu University of Technology (Shandong Academy of Sciences)\\
    \textsuperscript{\rm 2}Dalian Maritime University\\
    \textsuperscript{\rm 3}Dalian Polytechnic University\\
}

\begin{document}

\maketitle

\begin{abstract}
 Existing watermark attacks typically rely on predefined signal-processing operations or locally constrained restoration networks, making it difficult to capture the long-range dependencies of globally distributed watermark signals and resulting in an unfavorable trade-off between removal effectiveness and visual fidelity. In this paper, we propose \textit{SPFM-Net}, a semantic-prior-guided and frequency-constrained Mamba framework for invisible watermark attack. \textit{SPFM-Net} first employs high-ratio masking to disrupt the spatial coherence of invisible watermark signals, and then utilizes a partially fine-tuned pretrained Masked Autoencoder to reconstruct semantically consistent image from sparse observations while suppressing watermark-related information. A Multi-scale Residual Frequency Feature Interaction module subsequently aggregates watermark-related residual features across multiple receptive fields, while adaptively suppressing responses from watermark-irrelevant regions. To further capture the long-range dependencies of globally distributed watermark signals, a lightweight Mamba-based Global State-space Feature Modeling (GSFM) unit is introduced to separate watermark-related features from natural image content and suppress the remaining watermark traces. In addition, \textit{SPFM-Net} is optimized using a multi-level objective that jointly imposes spatial-, frequency-, and edge-domain constraints, enabling effective watermark suppression while preserving perceptual quality. Extensive experiments on representative spatial-domain, transform-domain, orthogonal moment-based, and deep learning-based watermarking schemes demonstrate that \textit{SPFM-Net} achieves a favorable trade-off between watermark attack effectiveness and perceptual fidelity.
\end{abstract}

\section{Introduction}

Invisible image watermarking embeds imperceptible information into visual content and has become an important technique for copyright protection, ownership verification, and content authentication \cite{ref_33,ref_34,ref_35}. A practical watermarking method should not only maintain visual quality but also ensure reliable watermark extraction under routine post-processing and adversarial manipulation \cite{ref_36}. Therefore, studying invisible watermark attack is essential for evaluating the security boundaries of existing watermarking methods \cite{ref_14,ref_15,ref_16,ref_17,ref_18,ref_19,ref_20,ref_21,ref_22,ref_23}. A major problem in watermark attacks is balancing watermark suppression and image fidelity: stronger modifications improve removal but degrade image quality, while weaker modifications preserve visual imperceptibility but may leave the watermark recoverable.

Existing watermark attacks, ranging from conventional image transformations to CNN-based and generative approaches, predominantly formulate watermark removal as a local perturbation or pixel-level restoration task. Such formulations have limited capability to distinguish watermark-related signals from reconstructed details, particularly when watermark information is deeply coupled with semantic and frequency representations. From an image restoration perspective, invisible watermarks can be viewed as weak but structured perturbations superimposed on the underlying image content, inducing subtle deviations in local statistics and feature representations while preserving perceptual imperceptibility. Therefore, effective attack requires more than simply suppressing pixel-level perturbations; it should recover intrinsic image representations while selectively weakening watermark-related information. Existing approaches remain limited in this regard for three main reasons. (1) \textbf{insufficient semantic guidance}: the lack of reliable semantic priors makes it difficult to distinguish watermark-related features from image content; (2) \textbf{limited global dependency modeling}: architectures dominated by local feature extraction are constrained by receptive fields, making it difficult to model long-range dependencies among spatially distributed watermark features; (3) \textbf{removal--fidelity trade-off}: stronger watermark attack usually requires more modification of high-frequency regions, leading to texture degradation and visible artifacts.

To this end, we propose \textit{SPFM-Net}, a semantic-prior-guided frequency-constrained mamba for universal invisible watermark attack. Instead of directly learning to remove watermark signals, \textit{SPFM-Net} reformulates watermark attack as a semantic-guided image restoration problem, aiming to recover intrinsic image content while suppressing watermark-related features. Specifically, \textit{SPFM-Net} first leverages a pre-trained masked autoencoder to derive reliable semantic priors from partially observed image content, thereby reducing the interference of watermark-related local details. Guided by these priors, a Multi-scale Residual Frequency Feature Interaction (MRFFI) module captures watermark-related discrepancies across different spatial scales and frequency components. A Mamba-based Global State-space Feature Modeling (GSFM) module is further introduced to model long-range dependencies among spatially distributed watermark features. Finally, joint pixel-, frequency-, and structure-level constraints further balance watermark removal and visual fidelity. Our main contributions are summarized as follows.

\begin{enumerate}
    \item We propose \textit{SPFM-Net}, a semantic-prior-guided frequency-constrained framework that reformulates invisible watermark removal as a semantic-guided image restoration task rather than direct perturbation suppression.
    \item We introduce a semantic-prior-guided frequency modeling strategy, where semantic priors facilitate content--watermark separation and MRFFI captures multi-scale frequency discrepancies for watermark suppression with reduced visual distortion.
    \item We design a Mamba-based GSFM module to capture long-range watermark dependencies, while multi-domain constraints preserve high-frequency details and improve the watermark removal--fidelity trade-off.
    \item Extensive experiments demonstrate strong attack effectiveness and generalization, with \textit{SPFM-Net} achieving zero-shot removal on unseen deep watermarks without target-specific fine-tuning, validating the feasibility of semantic-guided image reconstruction.
\end{enumerate}
\section{Related Work}
\subsection{Invisible Watermarking}
Existing watermarking methods present distinct strengths and vulnerabilities across different domains. Traditional spatial techniques (e.g., LSB), transform (e.g., DFT, DCT, wavelet), or orthogonal moment domains (e.g., PHFMs, RHFMs). While spatial methods offer high fidelity, they lack robustness against common processing. Transform and orthogonal moment methods improve resistance against specific distortions like rotation and scaling, yet remain sensitive to complex geometric changes or non-stationary noise, often incurring high computational costs. Conversely, end-to-end neural watermarking frameworks \cite{ref_39} provide stronger imperceptibility and robustness. For example, HiDDeN \cite{ref_1} introduced an encoder-decoder with a differentiable noise layer, and Jia \cite{ref_2} incorporated real and simulated JPEG compression during training. Yu \cite{ref_3} proposed adversarial watermark generation, while RAIMARK \cite{ref_4,ref_37} utilized implicit neural representations to enhance cross-resolution robustness. By implicitly encoding information into high-dimensional semantic and frequency representations, these advanced methods significantly complicate watermark removal.
\subsection{Invisible Watermark Attack}
Watermark attacks aim to disrupt embedded signals while preserving visual quality relative to the watermarked images. Early traditional signal-processing attacks (e.g., Gaussian noise, filtering, JPEG compression) effectively damage spatial-domain watermarks but cause severe image degradation against robust schemes. Consequently, deep learning-based attacks explore more effective removal. GAN-based methods, such as RWCA \cite{ref_29} and perceptual loss-guided attacks \cite{ref_30}, reconstruct images via adversarial learning but may introduce local artifacts. CNN-based denoising approaches, including HIWANet \cite{ref_5} and WAN \cite{ref_6}, suffer from limited receptive fields, struggling to remove globally distributed watermark patterns. Although diffusion-based methods like DDWRM \cite{ref_7,ref_42} and WARDM \cite{ref_28} achieve stronger removal through iterative restoration, their high computational costs and insufficient control over high-frequency residues frequently cause perceptual distortions measured by LPIPS\cite{ref_27} and A-FINE \cite{ref_32}. Furthermore, recent studies increasingly expose the instability of latent diffusion watermarks like Stable Signature \cite{ref_21} and exploring realistic removal strategies under restrictive no-box settings \cite{ref_31,ref_38}.
\subsection{Masked Autoencoders(MAE)}
Balancing removal effectiveness, visual fidelity, and computational efficiency remains a critical challenge. Masked Autoencoders (MAE) \cite{ref_12,ref_40} learn strong semantic priors by reconstructing heavily masked patches, capturing the global natural image structure while resisting local high-frequency disturbances. Although highly suitable for image restoration, MAE's semantic reconstruction capability remains underexplored for watermark attacks. Motivated by this, \textit{SPFM-Net} introduces a pretrained MAE encoder as the semantic backbone. By combining these global semantic priors with multi-scale frequency modeling, our method effectively separates watermark features from natural image content\cite{ref_41}, achieving thorough watermark removal while preserving high visual fidelity.
\section{Method}
\begin{figure*}[t]
    \centering
    \includegraphics[width=\textwidth]{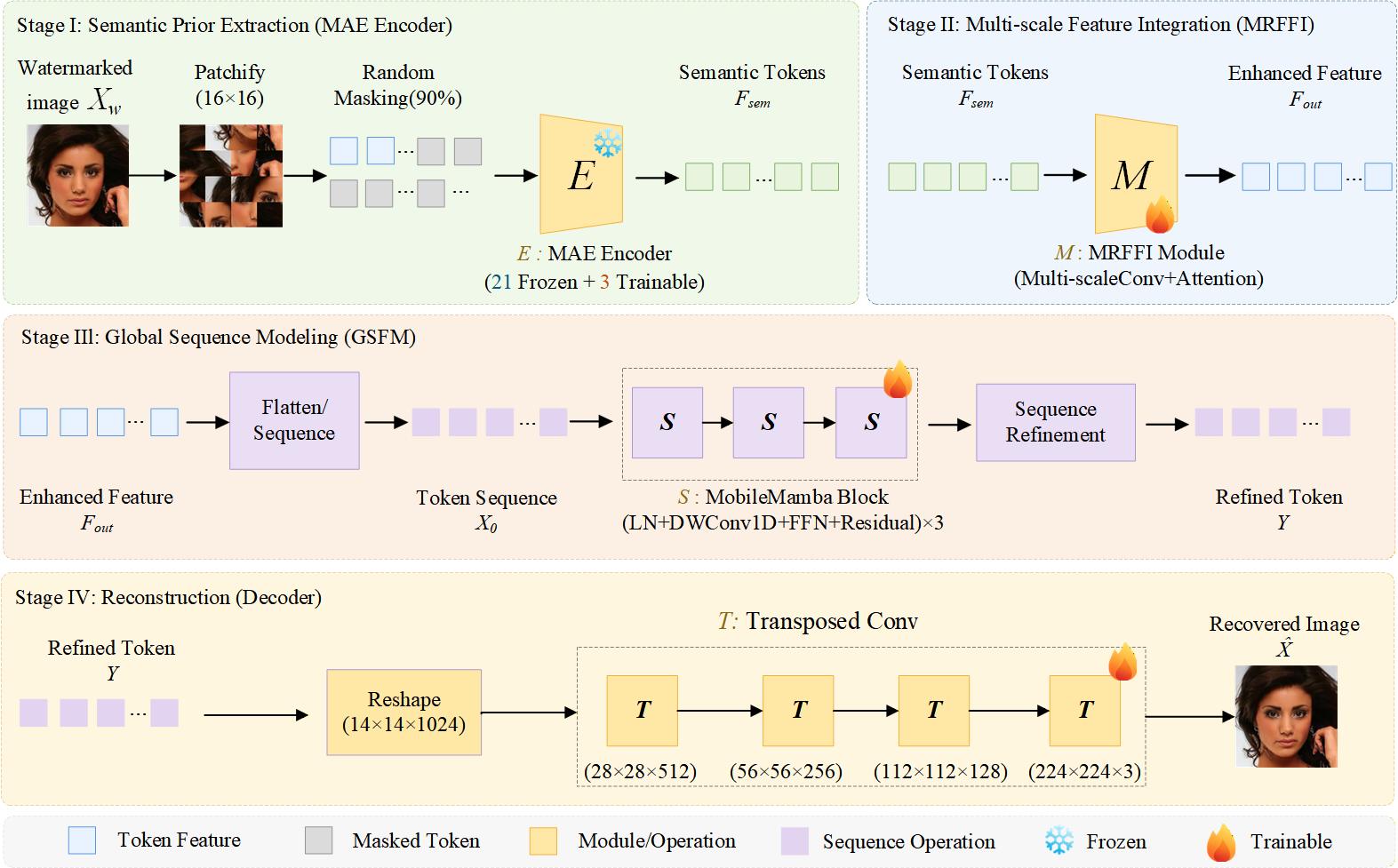}
    \caption{Overview of the proposed SPFM-Net.}
    \label{fig:overview}
\end{figure*}
\subsection{Overview}
As illustrated in Figure. 1, the proposed \textit{SPFM-Net} removes invisible watermarks through an image restoration pipeline guided by high-level semantic priors. Given a single watermarked image $X_w$, the input is first patchified and randomly masked before being fed into the pretrained MAE encoder $E$. The encoder extracts semantic features $F_{sem}$ while preserving the global structure of natural images. To adaptively fit faint watermark perturbations, the encoder remains mostly frozen, with only the final three blocks unfrozen for optimization. Subsequently, $F_{sem}$ is processed by the MRFFI module $M$ to capture multi-scale watermark residuals and suppress irrelevant responses through adaptive attention. The refined features are then fed into the Mamba-based GSFM module $S$, which models the long-range dependencies of globallydistributed watermark patterns and separates residual watermark information from natural image content. Finally, the decoder T reconstructs the watermark-free image $\hat{X}$ under spatial-frequency joint constraints with skip connections. 

\subsection{Image Patching and Random Masking}

Modern deep learning watermarks implicitly embed copyright information into spatial or high-dimensional features, deeply entangling with natural textures and complicating segmentation. Processing these images as continuous entities through conventional convolutional or self-attention modules often traps networks in local smoothing, confusing faint watermarks with host textures and causing incomplete removal. To sever the topological coherence of watermark signals without degrading natural high-frequency edges, we introduce an image patching and random masking mechanism at the pipeline frontend.
Given an input $X \in \mathbb{R}^{H \times W \times C}$, the network divides it into non-overlapping patches $X_p = \{X_p^{(1)}, X_p^{(2)}, \dots, X_p^{(N)}\} \in \mathbb{R}^{N \times (P^2C)}$. Here, $P=16$ represents the patch size, and $N=HW/P^2$ denotes the flattened sequence length. A uniform random masking strategy is then applied. We define a randomly sampled mask index set $\mathcal{F}$ controlled by a mask ratio $\rho$ set to 0.9. By discarding patches in $\mathcal{F}$ (where $\mathcal{F} \subset \{1,2,\dots,N\}$ such that $|\mathcal{F}| = \rho \times N$), the network retains only a sparse sequence of visible patches $X_v = \{X_p^{(i)} \mid i \notin \mathcal{F}\} \in \mathbb{R}^{N(1-\rho) \times (P^2C)}$. 
This design offers two advantages. First, extensive spatial masking destroys the continuity and joint distribution of watermark signals, preventing local denoising shortcuts. Second, the extremely sparse input severs local pixel dependencies, forcing the model to abandon superficial detail reconstruction and leverage global contextual priors to deduce deep image semantics.
\subsection{Partially Unfrozen Masked Encoder}
We introduce a masked encoder $E$, pre-trained via self-supervision on large-scale natural scenes, as the primary feature backbone using a partial unfreezing strategy. As shown in Figure 2, the watermarked input $X_w \in \mathbb{R}^{224 \times 224 \times 3}$ is partitioned into non-overlapping patches, flattened, and linearly projected into tokens. These tokens receive fixed 2D sine-cosine positional embeddings, and a classification token is appended. The sequence then enters $E$, comprising 24 Transformer blocks. Each block contains a Multi-Head Self-Attention mechanism and a feed-forward network, with a hidden dimension of 1024 and 16 attention heads. This high-dimensional extraction is expressed as:
\begin{equation}
T_{global} = E(X_w) \in \mathbb{R}^{N \times C}
\end{equation}
where $N=(H/P) \times (W/P)=14 \times 14=196$ denotes the sequence length. 
\begin{figure}[t]
    \centering
    \includegraphics[width=\columnwidth]{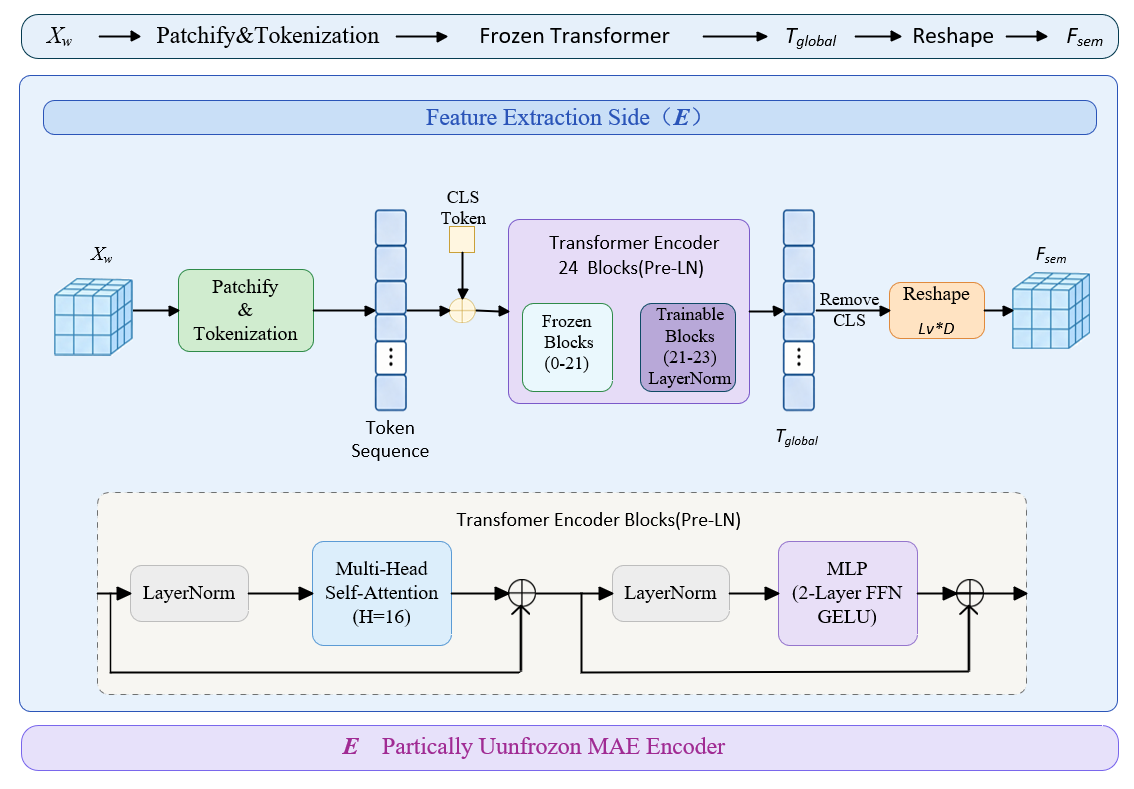} 
    \caption{Partially Unfrozen Masked Encoder Architecture.}
    \label{fig:my_single_column_fig}
\end{figure}

To adapt the pre-trained model for invisible watermark removal, we apply a local unfreezing strategy during joint optimization. Specifically, the first 21 Transformer blocks remain frozen to preserve macro-level structures and semantic priors, preventing quality degradation. Conversely, the final three blocks (blocks.21, blocks.22, and blocks.23) and the normalization layer are explicitly unfrozen for gradient backpropagation. These trainable components capture and adapt to subtle watermark residuals in the high-level feature space. Finally, after discarding the CLS token, the 1D sequence feature $T_{global}$ is reshaped and transposed to recover the spatial representation required downstream, formulated as $F_{sem} \in \mathbb{R}^{1024 \times 14 \times 14}$.

\subsection{MRFFI Module}

The MRFFI module utilizes parallel branches with diverse receptive fields to collaboratively extract multi-granularity watermark residuals, effectively overcoming the limitations of single-receptive-field mechanisms that struggle to balance capturing globally continuous perturbations and preserving natural edges across complex energy distributions.\cite{ref_11}.

As shown in Figure 3, given $F_{sem} \in \mathbb{R}^{1024 \times 14 \times 14}$ from $E$, $M$ utilizes three parallel $3 \times 3$ convolution branches with distinct strides ($s \in \{1, 2, 4\}$) to capture watermark features from local textures, intermediate structures, and global semantics:
\begin{align}
F_1 &= \text{Conv}_{3\times3}^{s=1}(F_{sem}) \in \mathbb{R}^{256 \times 14 \times 14} \\
F_2 &= \text{Interp}(\text{Conv}_{3\times3}^{s=2}(F_{sem})) \in \mathbb{R}^{256 \times 14 \times 14} \\
F_3 &= \text{Interp}(\text{Conv}_{3\times3}^{s=4}(F_{sem})) \in \mathbb{R}^{256 \times 14 \times 14}
\end{align}

\begin{figure}[t]
    \centering
    \includegraphics[width=\columnwidth]{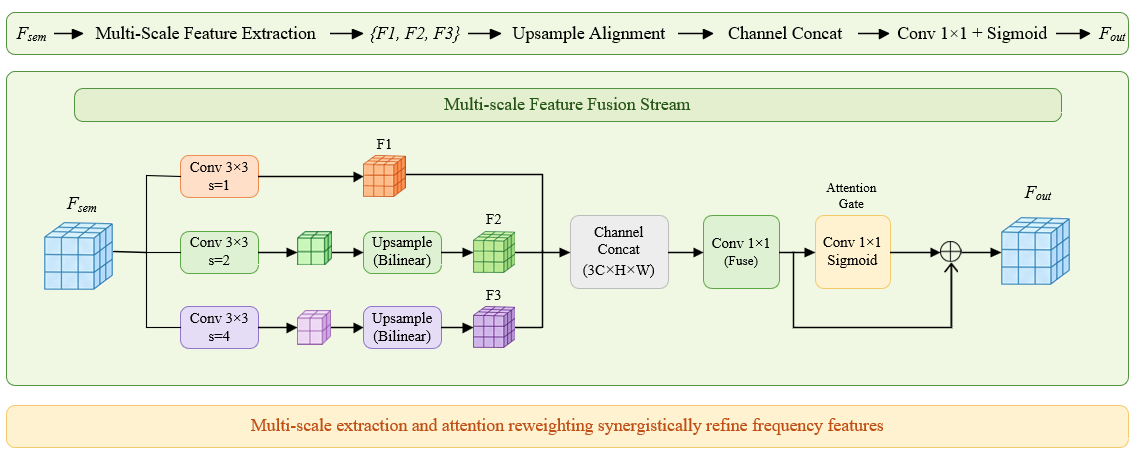} 
    \caption{MRFFI Module Architecture.}
    \label{fig:my_single_column_fig}
\end{figure}
All branches output 256 channels using SAME padding. For strides $s \in \{2, 4\}$, bilinear interpolation $\text{Interp}(\cdot)$ spatially aligns the down-sampled resolutions to $14 \times 14$. This yields three feature representations with identical resolutions encoding different information levels: $F_1$, $F_2$, and $F_3$.

To promote collaboration across receptive fields, MRFFI concatenates the features along the channel dimension and employs a $1 \times 1$ convolution for nonlinear interaction and dimensional alignment: $f_c = \text{Concat}(F_1, F_2, F_3) \in \mathbb{R}^{768 \times 14 \times 14}$ and $f_m = \text{Conv}_{1 \times 1}(f_c) \in \mathbb{R}^{1024 \times 14 \times 14}$. Since watermark perturbations distribute unevenly, uniform feature processing risks damaging clean backgrounds. Thus, $M$ introduces a channel attention mechanism to generate an adaptive weight map $A$:
\begin{align}
A &= \sigma(\text{Conv}_{1\times1}(f_m)) \\
F_{out} &= A \odot f_m \in \mathbb{R}^{1024 \times 14 \times 14}
\end{align} 
where $\sigma(\cdot)$ is the Sigmoid function and $\odot$ denotes the Hadamard product. This dynamic weighting significantly enhances watermark-related region responses while suppressing clean background interference, yielding the purified feature $F_{out}$.
\subsection{Mamba-based GSFM Module}
To enhance residual watermark removal, we propose the lightweight Mamba-based Global State-space Feature Modeling (GSFM) module. As shown in Figure 4, GSFM comprises three cascaded dual-residual sequence modeling blocks. The feature $F_{out}$ from MRFFI is reshaped into a 1D token sequence: $X = \text{Flatten}(F_{out}) \in \mathbb{R}^{196 \times 1024}$. Each block uses 1D depthwise separable convolution for efficient modeling. The forward refinement is formulated as: 
\begin{equation}
X' = X + \text{DWConv1D}(\text{LN}(X))
\end{equation} 
where $\text{LN}(\cdot)$ denotes layer normalization, and $\text{DWConv1D}(\cdot)$ is a depthwise convolution with kernel size 3, enhancing local contextual interaction among neighboring tokens with low overhead. 
Subsequently, a feed-forward transformation refines discriminative watermark representations:
\begin{equation}
Y = X' + \text{FFN}(\text{LN}(X'))
\end{equation}
where $\text{FFN}(\cdot)$ contains two linear projection layers with an expansion ratio of 2 and a GELU activation. 

This dual-residual design preserves semantics while progressively enhancing watermark-sensitive features. GSFM's global sequence scanning accurately isolates image-wide watermark traces from incomplete sequences, outputting a highly discriminative representation $Y \in \mathbb{R}^{196 \times 1024}$.
\begin{figure}[t]
    \centering
    \includegraphics[width=\columnwidth]{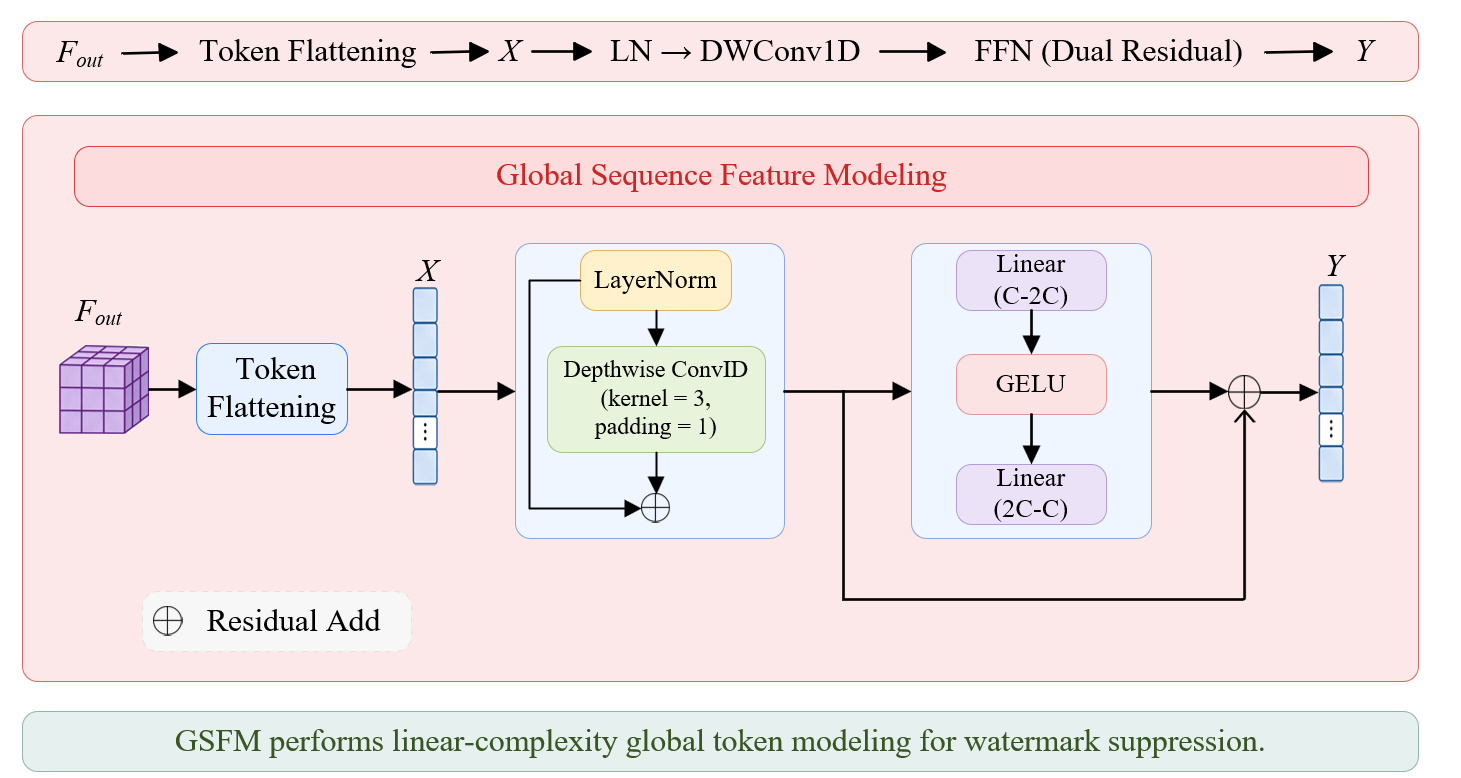} 
    \caption{Mamba-based GSFM Module Architecture.}
    \label{fig:my_single_column_fig}
\end{figure}
\subsection{Decoder and Optimization}
\textbf{Decoder.}
To project the enhanced sequence $Y$ back into continuous pixel space, we design a lightweight decoder $T$ with a resolution restoration symmetric to the front-end encoder $E$. First, $Y$ is reshaped into a low-resolution spatial feature: $f_{dec} \in \mathbb{R}^{1024 \times 14 \times 14}$. $T$ contains four consecutive transposed convolution layers, each using a $2 \times 2$ kernel and a stride of 2. Feature channels are hierarchically reduced from 1024 to 512, 256, 128, and finally 3, while spatial resolution exponentially enlarges from 14 to 28, 56, 112, and 224 to progressively refine image details. 

Since feature reduction inevitably degrades structures and edge details, we introduce cross-level skip connections during reconstruction. This transfers high-resolution details from shallow MAE encoder layers to concatenate directly with corresponding decoder features, effectively alleviating blurring and structural distortion. Finally, a Sigmoid activation normalizes output pixels to [0,1], yielding a watermark-free restored image with original spatial dimensions: $\hat{X}= \text{Sigmoid}(T(f_{dec}))$

\textbf{Joint Loss Function.}
To guarantee visual fidelity and explicitly reinforce the destruction of watermarks concealed in frequency domains and edge details, we construct a multi-level joint loss function. 
The spatial pixel reconstruction loss uses the $L_1$ norm to penalize the absolute difference between the reconstructed image $x$ and the clean ground-truth $X_{clean}$, macroscopically ensuring tonal consistency: 
\begin{equation}
\mathcal{L}_{pixel} = \frac{1}{3HW} \sum_{c=1}^3 \sum_{i=1}^H \sum_{j=1}^W |x(i,j,c) - X_{clean}(i,j,c)|
\end{equation}

Because invisible watermarks distribute energy across specific frequency coefficients, pixel-level loss exhibits limited frequency sensitivity. Thus, we introduce a 2D Fast Fourier Transform (FFT) constraint. Minimizing amplitude differences between predicted and ground-truth images encourages the network to reduce abnormal watermark-induced amplitude variations: 
\begin{equation}
\mathcal{L}_{fft} = \|\mathcal{F}_{2D}(x) - \mathcal{F}_{2D}(X_{clean})\|_1
\end{equation}
where $\mathcal{F}_{2D}(\cdot)$ denotes the spatial 2D FFT, and $\|\cdot\|_1$ is the $L_1$ norm of the magnitude difference. To suppress boundary artifacts, a spatial topological edge constraint utilizes the Sobel operator to extract horizontal and vertical gradient intensities: 
$\text{Sobel}(X) = \sqrt{\nabla_x(X_{gray})^2 + \nabla_y(X_{gray})^2 + \epsilon_0}$ 
\begin{equation}
\mathcal{L}_{edge} = \|\text{Sobel}(x) - \text{Sobel}(X_{clean})\|_1
\end{equation}
where $X_{gray}$ is the grayscale image and $\epsilon_0=10^{-6}$ is a smoothing constant. This aligns edge gradient distributions. Finally, based on empirical optimization, the three loss terms construct the joint objective function: 
\begin{equation}
\mathcal{L}_{total} = \mathcal{L}_{pixel} + \lambda_1 \mathcal{L}_{fft} + \lambda_2 \mathcal{L}_{edge}
\end{equation}
with hyperparameters set to $\lambda_1 = 0.5$ and $\lambda_2 = 0.3$. This joint constraint guarantees both high-fidelity reconstruction and effective watermark removal.

\section{Experiments}
\subsection{Experimental Settings}
\begin{table*}[t]
    \centering
    \renewcommand{\arraystretch}{1.15} 
    \begin{tabular}{l ccccccc | c}
        \hline
        \textbf{Attacks} & \textbf{JPEG} & \textbf{Crop} & \textbf{Blur} & \textbf{Noise} & \textbf{DiffusionAttack} & \textbf{VAEAttack} & \textbf{UnMarker} & \textit{\textbf{SPFM-Net}} \\
        \hline
        SS     & 0.06 & 0    & 0.08 & 0.06 & 0.44 & 0.46 & \textbf{0.48} & \textbf{0.48} \\
        Yu     & 0.20 & 0.23 & 0.05 & 0.30 & 0.38 & \textbf{0.41} & 0.37 & 0.12 \\
        Hidden & 0.30 & 0.10 & 0.33 & 0.43 & 0.43 & 0.43 & 0.50 & \textbf{0.55} \\
        \hline
    \end{tabular}
    \caption{Comparison of BER performance among different attack under three deep learning-based watermark schemes.}
\end{table*}

\begin{table*}[t]
    \centering
    \renewcommand{\arraystretch}{1.15} 
    
    \resizebox{\textwidth}{!}{
    \begin{tabular}{l cccc cccc cccc}
        \hline
        \multicolumn{1}{c}{\multirow{2}{*}{\textbf{WM}}} & \multicolumn{4}{c}{\textbf{SS}} & \multicolumn{4}{c}{\textbf{Yu}} & \multicolumn{4}{c}{\textbf{Hidden}} \\
        \cline{2-5} \cline{6-9} \cline{10-13}
        & PSNR & SSIM & LPIPS & A-FINE & PSNR & SSIM & LPIPS & A-FINE & PSNR & SSIM & LPIPS & A-FINE \\
        \hline
        JPEG & 28.54 & 0.98 & 0.05 & 5.70 & 7.01 & 0.01 & 0.44 & 69.99 & 33.55 & \textbf{0.99} & \textbf{0.01} & 48.14 \\
        Crop & 12.95 & 0.64 & 0.28 & 54.16 & 7.17 & 0.02 & 0.45 & 76.45 & 16.31 & 0.74 & 0.11 & 55.60 \\
        Blur & 27.89 & 0.98 & 0.17 & 0.61 & 7.04 & 0.01 & 0.51 & 83.12 & 29.92 & 0.98 & 0.13 & 76.47 \\
        Noise & 24.14 & 0.97 & 0.26 & 52.56 & 6.98 & 0.01 & 0.53 & 69.86 & 25.91 & 0.96 & 0.14 & 54.04 \\
        \hline
        DiffusionAttack & 24.90 & 0.57 & 0.34 & \textbf{0.10} & 18.37 & 0.83 & 0.10 & 60.37 & 29.81 & 0.74 & 0.12 & 64.80 \\
        VAEAttack & 26.98 & 0.98 & 0.24 & 19.02 & 30.43 & \textbf{0.99} & 0.17 & 75.66 & 27.72 & 0.97 & 0.16 & 77.63 \\
        UnMarker & 22.22 & 0.55 & 0.29 & 72.76 & 18.23 & 0.88 & \textbf{0.08} & 50.97 & 25.03 & 0.71 & 0.11 & 57.85 \\
        \hline
        \textit{SPFM-Net} & \textbf{32.83} & \textbf{0.99} & \textbf{0.04} & 50.47 & \textbf{42.76} & \textbf{0.99} & 0.13 & \textbf{26.86} & \textbf{40.65} & \textbf{0.99} & 0.07 & \textbf{30.36} \\
        \hline
    \end{tabular}
    } 
    \caption{Comparison of Visual Quality among Various invisible watermark attacks.}
\end{table*}
\textbf{Implementation Details and Training.} 
We implement \textit{SPFM-Net} using PyTorch 2.1.0 on a single NVIDIA RTX 4090 GPU. The visual backbone leverages self-supervised pre-trained weights (\texttt{checkpoint-300.pth}). During joint fine-tuning, the first 21 Transformer blocks are frozen; only bocks 21, 22, and 23 are updated to specifically fit faint watermark noise features. We train for 100 epochs using the AdamW optimizer (learning rate $1 \times 10^{-5}$, weight decay $1 \times 10^{-6}$) and a batch size of 2. To enhance robustness, random Gaussian noise ($\sigma=0.02$) is superimposed during forward propagation. The optimization objective is a weighted combination of the spatial pixel reconstruction loss, 2D-FFT constraint, and Sobel topological edge loss. 

\textbf{Datasets and Watermarking Methods.}
All images are resized to $224 \times 224$. For the 20,000-image training set, we embed multi-scale watermarks ($16 \times 16$ and $32 \times 32$) into internally loaded clean face images using traditional schemes (LSB, DCT, QPHFMs). For evaluation, we randomly sample 100 images per advanced deep watermarking target: CelebA-HQ for Yu, Stable Diffusion Prompts (SDP) for Stable Signature (SS), and SDXL 1.0 (Base) for HiDDeN. Notably, the evaluated deep learning-based schemes are absent from the training phase. This cross-mechanism setup allows us to objectively assess the zero-shot generalization capability of our semantic-prior-guided attack.

\textbf{Baseline Attacks and Metrics.} 
We benchmark \textit{SPFM-Net} against conventional signal processing distortions (JPEG $QF=80$, cropping ratio $0.9$, Gaussian blur, and Gaussian noise $\sigma^2=0.05$) and representative SOTA deep learning attacks (DiffusionAttack \cite{ref_24}, VAEAttack \cite{ref_25}, and UnMarker \cite{ref_26}). We use Bit Error Rate (BER) to quantify watermark destruction, where values approaching 0.5 indicate random extraction (strong attack). To evaluate visual fidelity, we report PSNR, SSIM, LPIPS, and A-FINE. Crucially, all image quality metrics are strictly computed in a paired manner between the attacked images and their corresponding watermarked images.
\subsection{Attack Effectiveness Evaluation}
\textbf{Attack Effectiveness.} 
Table 1 demonstrates that traditional signal processing attacks struggle against modern deep watermarks (e.g., spatial cropping on SS yields a 0 BER, indicating 100\% recovery). Conversely, \textit{SPFM-Net} achieves significant effectiveness against advanced deep watermarking schemes. On SS and HiDDeN, \textit{SPFM-Net} yields BERs of 0.48 and 0.55, respectively, which are comparable to or higher than those of the unconstrained UnMarker (0.48 and 0.50) and closely approach the binary random guessing level. This confirms that integrating multi-scale feature extraction and global sequence scanning effectively models and neutralizes globally distributed watermark perturbations.

\textbf{Zero-shot Cross-Mechanism Generalization.} 
Table 1 highlights \textit{SPFM-Net}'s zero-shot capability. Although trained exclusively on traditional watermarks (LSB, DCT, QPHFMs), the network generalizes to disrupt unseen deep generative watermarks, such as SS (BER 0.48) and HiDDeN (BER 0.55). This stems from our semantic-prior-guided formulation; instead of overfitting to specific noise patterns, the frozen MAE encoder forces reconstruction to strictly follow natural image priors. Consequently, embedded watermarks, regardless of their generation mechanisms, are treated as high-frequency anomalies and filtered out during decoding.

\textbf{Limitation.} 
However, this strong prior constraint introduces a trade-off on specific spatial-domain schemes like Yu, yielding a lower BER (0.12) compared to the unconstrained UnMarker (0.37). By restricting the degradation of edge textures to prioritize high fidelity, a small amount of watermark signals deeply coupled with high-frequency details may survive. Nevertheless, this conservative strategy guarantees a significant advantage in post-attack visual quality (42.76 dB PSNR vs. UnMarker's 18.23 dB), presenting a practical compromise for realistic scenarios.
\subsection{Visual Quality Comparison of Attacked Images}
\textbf{Quantitative Evaluation.}
As detailed in Table 2,\textit{SPFM-Net} maintains high reconstruction fidelity. Existing deep learning attacks often sacrifice image quality for watermark disruption. For instance, UnMarker's PSNR drops to 22.22 dB on SS with an increase in LPIPS, while VAEAttack suffers perceptual degradation on Yu due to absent manifold constraints, causing high-frequency detail filtering. In contrast,\textit{SPFM-Net} achieves consistent performance across metrics, maintaining 0.99 SSIM and comparatively low LPIPS across SS and HiDDeN. This is attributed to robust semantic priors from the frozen encoder and edge gradient constraints that restrict pixel contour distortion.
\begin{table*}[t]
    \centering
    \renewcommand{\arraystretch}{1.15} 
    \begin{tabular*}{\textwidth}{@{\extracolsep{\fill}} l cccc cccc cccc @{}}
        \hline
        \multicolumn{1}{c}{\multirow{2}{*}{\textbf{WM}}} & \multicolumn{4}{c}{\textbf{SS}} & \multicolumn{4}{c}{\textbf{Yu}} & \multicolumn{4}{c}{\textbf{Hidden}} \\
        \cline{2-5} \cline{6-9} \cline{10-13}
        & PSNR & SSIM & LPIPS & A-FINE & PSNR & SSIM & LPIPS & A-FINE & PSNR & SSIM & LPIPS & A-FINE \\
        \hline
        \textit{Full} & 32.83 & \textbf{0.99} & \textbf{0.04} & 50.47 & \textbf{42.76} & \textbf{0.99} & \textbf{0.13} & \textbf{26.86} & \textbf{40.65} & \textbf{0.99} & \textbf{0.07} & \textbf{30.36} \\
        Baseline & 32.34 & \textbf{0.99} & 0.07 & 53.37 & 41.56 & 0.98 & 0.20 & 34.85 & 40.53 & \textbf{0.99} & 0.08 & 43.58 \\
        Variant 1 & 27.05 & 0.98 & 0.18 & 62.29 & 38.50 & \textbf{0.99} & 0.18 & 32.51 & 36.15 & \textbf{0.99} & 0.11 & 40.73 \\
        Variant 2 & \textbf{38.25} & \textbf{0.99} & 0.17 & \textbf{36.28} & 40.47 & \textbf{0.99} & 0.17 & 30.32 & 36.62 & \textbf{0.99} & 0.10 & 37.57 \\
        \hline
    \end{tabular*}
    \caption{Visual quality evaluation results of each variant across different watermarking scenarios.}
\end{table*}

\textbf{Visual Corroboration.}
Fig. 5 provides direct visual evidence of these quantitative results. VAEAttack induces noticeable over-smoothing, resulting in a reduction of facial textures, while UnMarker introduces color drift and irregular granular noise.\textit{SPFM-Net} effectively suppresses these artifacts, preserving structural integrity across both broad skin tone distributions and high-frequency regions (e.g., eyes). 
\begin{figure}[t]
    \centering
    \includegraphics[width=\columnwidth]{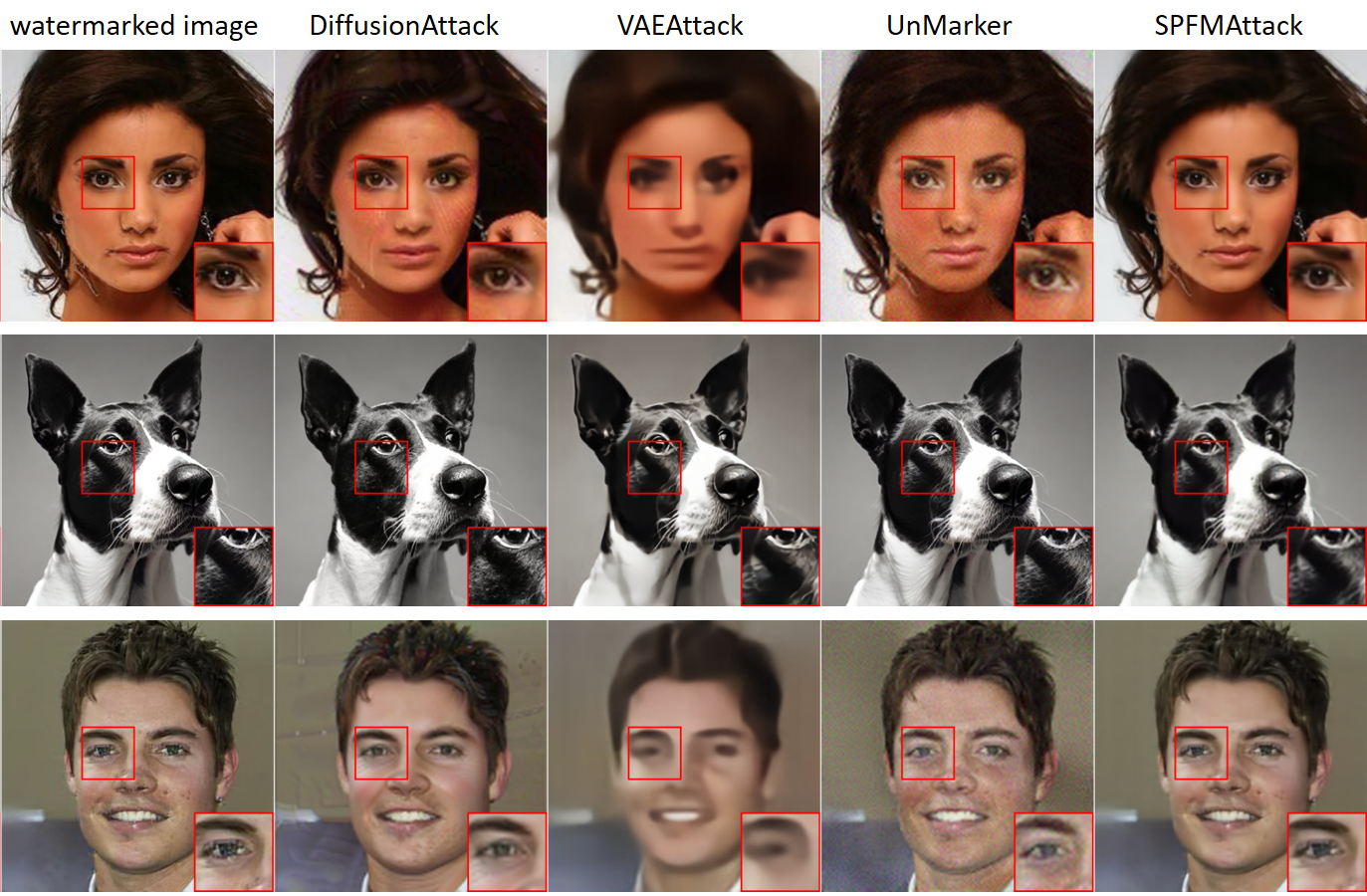} 
    \caption{SOTA removal attack outputs across different watermarking schemes. Top row: HiDDeN; Middle row: Stable Signature; Bottom row: LSB.}
    \label{fig:visual_comparison}
\end{figure}

\textbf{Performance Trade-off.}
Furthermore, the Pareto frontiers in Fig. 6 illustrate the two-dimensional distribution of visual fidelity (PSNR) and attack effectiveness (BER). Traditional methods are typically confined to the inefficient left region, while existing deep attacks, such as UnMarker and DiffusionAttack, fall into the low-fidelity bottom-right quadrant. Conversely, \textit{SPFM-Net} consistently occupies the top-right quadrant under both spatial (HiDDeN) and generative (SS) watermarking mechanisms, achieving a favorable balance between watermark disruption and visual fidelity.
\begin{figure}[t]
    \centering
    \includegraphics[width=\columnwidth]{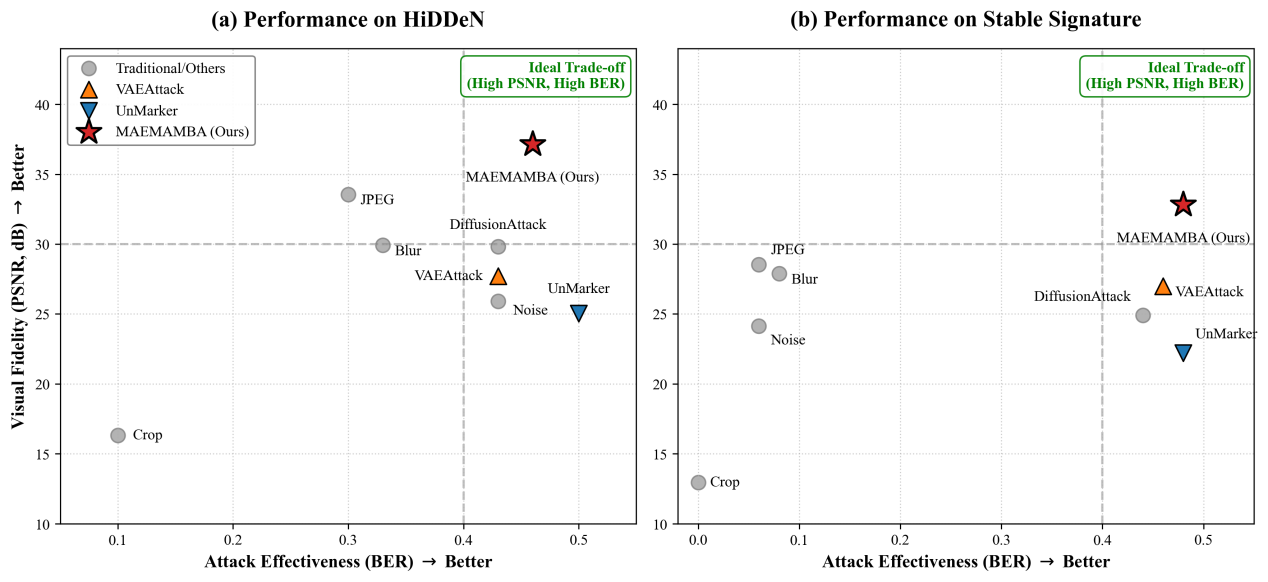} 
    \caption{Performance trade-off between attack effectiveness (BER) and visual fidelity (PSNR). (a) Spatial-domain watermark (HiDDeN). (b) Generative latent watermark (Stable Signature).}
    \label{fig:my_single_column_fig}
\end{figure}

\subsection{Ablation Study}
\textbf{Experimental Setup.}
We design progressive ablation studies to validate \textit{SPFM-Net}'s core components: the MRFFI module, GSFM module, and multi-level joint loss (Tables 3 and 4, Fig. 7). The baseline model comprises only the partially unfrozen masked encoder and foundational decoder. We sequentially introduce MRFFI (Variant 1), GSFM (Variant 2), and spatial-frequency composite constraints (Full).

\textbf{Module Contributions.}
The baseline model exhibits limited capability in disrupting complex deep watermarks. However, the progressive integration of the MRFFI and GSFM modules yields a steady increase in the extraction error. Notably, Variant 2 increases the BER to 0.52 on SS and 0.57 on HiDDeN, both of which surpass the theoretical expectation of binary random guessing ($\approx 0.5$). This substantial efficacy improvement substantiates the necessity of MRFFI and GSFM in modeling and neutralizing globally distributed hidden watermark features within high-dimensional spaces.
\begin{table}[t]
\centering
\renewcommand{\arraystretch}{1.2}
\begin{tabular*}{\columnwidth}{@{\extracolsep{\fill}}lccc@{}}
\hline
\multicolumn{1}{c}{\raisebox{-1.5ex}[0pt][0pt]{\textbf{WM}}} & \multicolumn{3}{c}{\textbf{BER}} \\
\cline{2-4}
& \textbf{Yu} & \textbf{SS} & \textbf{HiDDeN} \\
\hline
\textit{Full} & \textbf{0.12} & 0.48 & 0.55 \\
Baseline & 0.08 & 0.43 & 0.51 \\
Variant 1 & \textbf{0.12} & 0.47 & 0.54 \\
Variant 2 & \textbf{0.12} & \textbf{0.52} & \textbf{0.57} \\
\hline
\end{tabular*}
\caption{BER results of each variant across different watermarking scenarios.}
\end{table}

\textbf{Joint Constraints.}
As illustrated in Fig. 7, variants lacking composite loss constraints (Variants 1 and 2) experience PSNR reductions and perceptual degradation due to the excessive filtering of high-frequency features. Incorporating the 2D-FFT and Sobel edge losses (Full) improves visual fidelity (achieving 42.46 dB PSNR on Yu, the highest across variants) while maintaining disruption effectiveness. Ultimately, \textit{SPFM-Net} achieves a favorable trade-off, balancing watermark disruption via global correlation modeling and structural preservation via semantic priors and frequency-domain constraints.
\begin{figure}[t]
    \centering
    \includegraphics[width=\columnwidth]{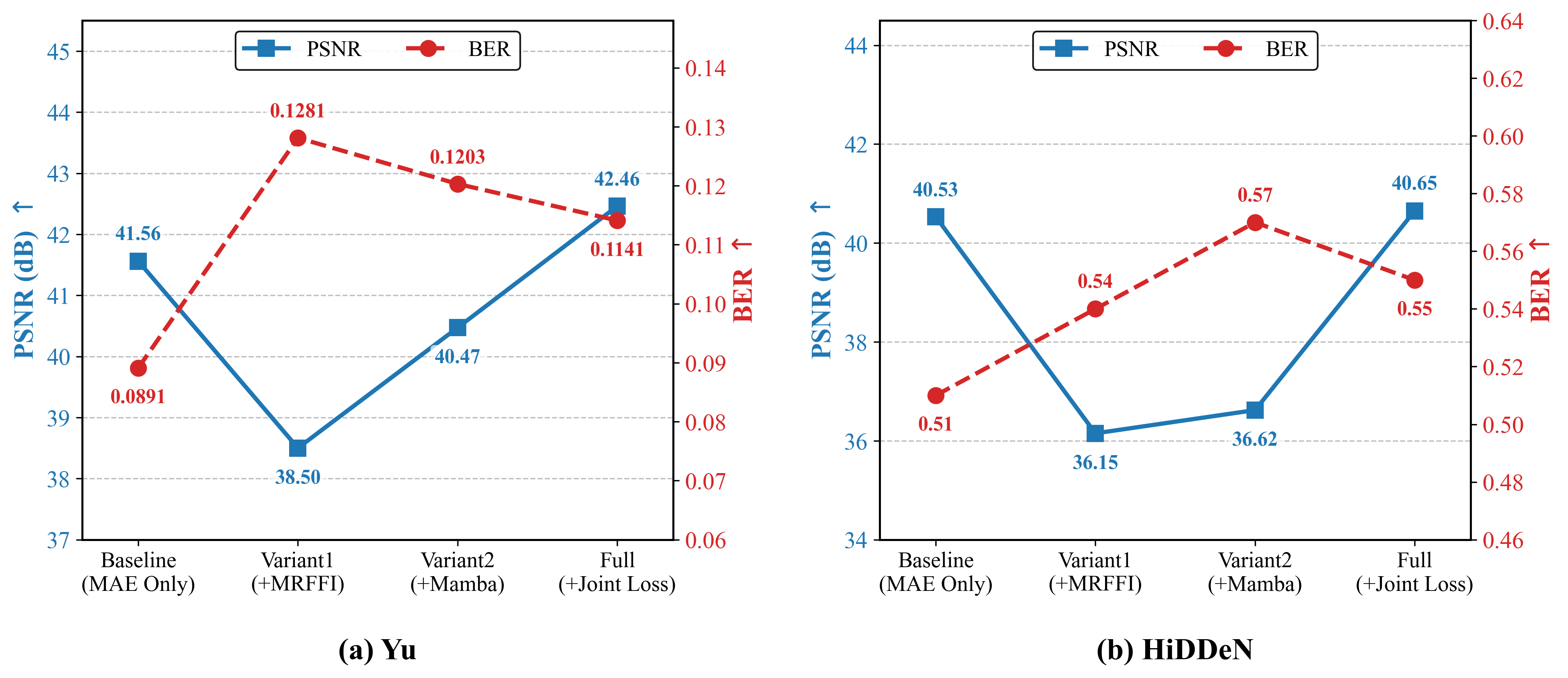} 
    \caption{Trade-off between image quality and attack effectiveness under HiDDeN and Yu watermark scenarios.}
    \label{fig:my_single_column_fig}
\end{figure}

\section{CONCLUSION}
This paper proposes \textit{SPFM-Net}, a universal image watermark attack network using semantic priors and frequency domain constraints to effectively remove traditional and deep learning invisible watermarks. Coupling extreme spatial masking with semantic-guided reconstruction, \textit{SPFM-Net} achieves invisible watermark removal while maintaining exceptionally high visual fidelity. Extensive experiments demonstrate that our method outperforms existing baselines in balancing attack effectiveness and image quality, offering a novel semantic decoupling perspective for digital watermarking. Given the rapid iteration of generative watermarking and adversarial steganography, future research will focus on enhancing the model's generalization stability against cross-modal scenarios and novel watermarks with dynamic topological changes to tackle broader applications.

\bibliography{aaai2027}

@inproceedings{ref_1,
  title={Hidden: Hiding data with deep networks},
  author={Zhu, Jiren and Kaplan, Russell and Johnson, Justin and Fei-Fei, Li},
  booktitle={Proceedings of the European Conference on Computer Vision (ECCV)},
  pages={657--672},
  year={2018}
}

@inproceedings{ref_2,
  title={MBRS: Enhancing Robustness of DNN-based Watermarking by Mini-Batch of Real and Simulated JPEG Compression},
  author={Jia, Z and Fang, H and Zhang, W},
  booktitle={ACM},
  pages={41--49},
  year={2021}
}

@inproceedings{ref_3,
  title={Responsible disclosure of generative models using scalable fingerprinting},
  author={Yu, Ning and Skripniuk, Vladislav and Chen, Dingfan and Davis, Larry S and Fritz, Mario},
  booktitle={International Conference on Learning Representations (ICLR)},
  year={2022}
}

@inproceedings{ref_4,
  title={Achieving resolution-agnostic dnn-based image watermarking: A novel perspective of implicit neural representation},
  author={Wang, Y and Zhu, X and Ye, G and others},
  booktitle={Proceedings of the 32nd ACM International Conference on Multimedia},
  pages={10354--10362},
  year={2024}
}

@article{ref_5,
  title={Highly applicable and imperceptible watermark attack network},
  author={Wang, Chunpeng and Tian, Pengfei and Xia, Zhiqiu and Li, Qi and Li, Jian and Wei, Ziqi and Luo, Ting and Ma, Bin},
  journal={Signal Processing},
  volume={230},
  pages={109840},
  year={2025},
  publisher={Elsevier}
}

@inproceedings{ref_6,
  title={WAN: Watermarking Attack Network},
  author={Nam, S H and Ahn, W and Yu, I J and others},
  booktitle={Proceedings of the 32nd British Machine Vision Conference (BMVC)},
  pages={1--13},
  year={2021},
  organization={BMVA Press}
}

@misc{ref_7,
  title={Diffusion Denoising Watermark Removal Models to Attack Invisible Image Watermarks},
  author={Mareen, H and De Meulenaere, K and Lambert, P and others},
  howpublished={Ghent University Academic Bibliography},
  year={2023}
}

@inproceedings{ref_11,
  title={SPFM-Net: Lightweight multi-receptive visual mamba network},
  author={He, H and Zhang, J and Cai, Y and others},
  booktitle={Proceedings of the IEEE/CVF Conference on Computer Vision and Pattern Recognition},
  pages={4497--4507},
  year={2025}
}

@inproceedings{ref_12,
  title={Masked autoencoders are scalable vision learners},
  author={He, K and Chen, X and Xie, S and others},
  booktitle={Proceedings of the IEEE/CVF Conference on Computer Vision and Pattern Recognition},
  pages={16000--16009},
  year={2022}
}

@article{ref_14,
  title={An imperceptible spatial domain color image watermarking scheme},
  author={Jobin, A. and Varghese, P.},
  journal={Journal of King Saud University - Computer and Information Sciences},
  volume={31},
  number={1},
  pages={125--133},
  year={2019},
  publisher={Elsevier}
}

@article{ref_15,
  title={Tamper localization and content restoration in fragile image watermarking: A review},
  author={Amrullah, A. and Aminuddin, A.},
  journal={International Journal of Advanced Computer Information},
  volume={2},
  pages={62--74},
  year={2026}
}

@article{ref_16,
  title={Robust image watermarking based on hybrid IWT-DCT-SVD},
  author={Wahyudi, M. I. and Fauzi, I. and Atmojo, D.},
  journal={International Journal of Advanced Computer Information},
  volume={1},
  number={2},
  pages={89--98},
  year={2025}
}

@article{ref_17,
  title={Robust digital image watermarking using DWT, Hessenberg, and SVD for copyright protection},
  author={Kusuma, M. R. and Panggabean, S.},
  journal={International Journal of Advanced Computer Information},
  volume={2},
  number={1},
  pages={41--52},
  year={2026}
}

@article{ref_18,
  title={Exploring accurate invariants on polar harmonic Fourier moments in polar coordinates for robust image watermarking},
  author={He, M. and Wang, H. and Zhang, F. and Xiang, Y.},
  journal={IEEE Transactions on Multimedia},
  volume={26},
  pages={5435--5449},
  year={2024},
  publisher={IEEE}
}

@article{ref_19,
  title={Fast image reconstruction method using radial harmonic Fourier moments and its application in digital watermarking},
  author={Zhang, H. and Li, Z. and Chen, Y. and Lu, C. and Yan, P.},
  journal={Journal of the Franklin Institute},
  volume={362},
  number={1},
  pages={107391},
  year={2025},
  publisher={Elsevier}
}

@article{ref_20,
  title={A two-stage robust reversible watermarking using polar harmonic transform for high robustness and capacity},
  author={Tang, Y. and Li, K. and Wang, C. and Bian, S. and Huang, Q.},
  journal={Information Sciences},
  volume={654},
  pages={119786},
  year={2024},
  publisher={Elsevier}
}

@misc{ref_21,
    title={Stable signature is unstable: removing image watermark from diffusion models},
    author={Hu, Y. and Jiang, Z. and Guo, M. and Gong, N.},
    year={2024},
    eprint={2405.07145},
    archivePrefix={arXiv}
}

@inproceedings{ref_22,
  title={Tree-rings watermarks: Invisible fingerprints for diffusion images},
  author={Wen, Yuxin and Kirchenbauer, John and Geiping, Jonas and Goldstein, Tom},
  booktitle={Advances in Neural Information Processing Systems (NeurIPS)},
  volume={36},
  pages={58047--58063},
  year={2023}
}

@inproceedings{ref_23,
  title={The stable signature: Rooting watermarks in latent diffusion models},
  author={Fernandez, Pierre and Couairon, Guillaume and J{\'e}gou, Herv{\'e} and Douze, Matthijs and Furon, Teddy},
  booktitle={Proceedings of the IEEE/CVF International Conference on Computer Vision (ICCV)},
  pages={22466--22477},
  year={2023}
}

@inproceedings{ref_24,
  title={Invisible image watermarks are provably removable using generative AI},
  author={Zhao, Xuandong and Zhang, Kexun and Su, Zihao and Vasan, Saastha and Grishchenko, Ilya and Kruegel, Christopher and Vigna, Giovanni and Wang, Yu-Xiang and Li, Lei},
  booktitle={Advances in Neural Information Processing Systems (NeurIPS)},
  volume={37},
  pages={8643--8672},
  year={2024}
}

@misc{ref_25,
  title={Waves: Benchmarking the robustness of image watermarks},
  author={An, Bang and Ding, Mucong and Rabbani, Tahseen and Agrawal, Aakriti and Xu, Yuancheng and Deng, Chenghao and Zhu, Sicheng and Mohamed, Abdirisak and Wen, Yuxin and Goldstein, Tom and others},
  howpublished={arXiv preprint arXiv:2401.08573},
  year={2024}
}

@inproceedings{ref_26,
  title={Unmarker: a universal attack on defensive image watermarking},
  author={Kassis, Andre and Hengartner, Urs},
  booktitle={Proceedings of the IEEE Symposium on Security and Privacy (SP)},
  pages={2602--2620},
  year={2025},
  organization={IEEE}
}

@inproceedings{ref_27,
  title={The unreasonable effectiveness of deep features as a perceptual metric},
  author={Zhang, R. and Isola, P. and Efros, A. A. and Shechtman, E. and Wang, O.},
  booktitle={Proceedings of the IEEE conference on computer vision and pattern recognition},
  pages={586--595},
  year={2018}
}

@article{ref_28,
  title={Can watermarks be removed like noise? A watermarking attack network using residual diffusion model},
  author={Wang, C. and Zhang, F. and Zhang, S. and others},
  journal={IEEE Transactions on Dependable and Secure Computing},
  year={2026},
  publisher={IEEE}
}

@inproceedings{ref_29,
  title={Detect and remove watermark in deep neural networks via generative adversarial networks},
  author={Sun, S. and Wang, H. and Xue, M. and others},
  booktitle={International Conference on Information Security},
  pages={341--357},
  year={2021},
  organization={Springer}
}

@article{ref_30,
  title={Concealed attack for robust watermarking based on generative model and perceptual loss},
  author={Li, Q. and Wang, X. and Ma, B. and others},
  journal={IEEE Transactions on Circuits and Systems for Video Technology},
  volume={32},
  number={8},
  pages={5695--5706},
  year={2021},
  publisher={IEEE}
}

@misc{ref_31,
    title={When there is no decoder: Removing watermarks from stable diffusion models in a no-box setting},
    author={Wu, X. and Tang, T. and Li, X. and Ni, J. and Yu, Y.},
    year={2025},
    eprint={2507.03646},
    archivePrefix={arXiv}
}

@inproceedings{ref_32,
  title={Toward generalized image quality assessment: Relaxing the perfect reference quality assumption},
  author={Chen, Du and Wu, Tianhe and Ma, Kede and Zhang, Lei},
  booktitle={Proceedings of the IEEE/CVF Conference on Computer Vision and Pattern Recognition (CVPR)},
  pages={12742--12752},
  year={2025}
}

@article{ref_33,
  title={A large-capacity and robust screen-shooting resilient image watermarking based on attention-enhanced invertible neural network},
  author={Zhang, J. and Zhang, Z. and Li, Z. and others},
  journal={Journal of Information Security and Applications},
  volume={97},
  pages={104346},
  year={2026},
  publisher={Elsevier}
}

@inproceedings{ref_34,
  title={Sleepermark: Towards robust watermark against fine-tuning text-to-image diffusion models},
  author={Wang, Z. and Guo, J. and Zhu, J. and others},
  booktitle={Proceedings of the Computer Vision and Pattern Recognition Conference},
  pages={8213--8224},
  year={2025}
}

@article{ref_35,
  title={Watermarking diffusion model},
  author={Liu, Y. and Li, Z. and Backes, M. and others},
  journal={arXiv preprint arXiv:2305.12502},
  year={2023}
}

@inproceedings{ref_36,
  title={TAG-WM: Tamper-Aware Generative Image Watermarking via Diffusion Inversion Sensitivity},
  author={Chen, Y. and Ma, Z. and Fang, H. and others},
  booktitle={Proceedings of the IEEE/CVF International Conference on Computer Vision},
  pages={16723--16732},
  year={2025}
}

@article{ref_37,
  title={TrustMark: Universal Watermarking for Arbitrary Resolution Images},
  author={Bui, T. and Agarwal, S. and Collomosse, J. P.},
  journal={arXiv preprint arXiv:2311.18297},
  year={2023}
}

@inproceedings{ref_38,
  title={Practical Deep Dispersed Watermarking with Synchronization and Fusion},
  author={Guo, H. and Zhang, Q. and Luo, J. and Guo, F. and Zhang, W. and Su, X. and Li, M.},
  booktitle={Proceedings of the ACM International Conference on Multimedia},
  pages={7922--7932},
  year={2023}
}

@inproceedings{ref_39,
  title={Towards Blind Watermarking: Combining Invertible and Non-invertible Mechanisms},
  author={Ma, R. and Guo, M. and Hou, Y. and Fan, Y. and Li, Y. and Jia, H. and Xie, X.},
  booktitle={Proceedings of the ACM International Conference on Multimedia},
  pages={1532--1542},
  year={2022}
}

@article{ref_40,
  title={A self-supervised network for image denoising and watermark removal},
  author={Tian, C. and Xiao, J. and Zhang, B. and Zuo, W. and Zhang, Y. and Lin, C.},
  journal={Neural Networks},
  volume={174},
  pages={106218},
  year={2024},
  publisher={Elsevier}
}

@inproceedings{ref_41,
  title={DENet: Disentangled embedding network for visible watermark removal},
  author={Sun, R. and Su, Y. and Wu, Q.},
  booktitle={Proceedings of the AAAI Conference on Artificial Intelligence},
  pages={2411--2419},
  year={2023}
}

@inproceedings{ref_42,
  title={DiffWA: Diffusion models for watermark attack},
  author={Li, X.},
  booktitle={International Conference on Intelligent Information and Computing Systems (ICIICS)},
  pages={1--8},
  year={2023}
}


\end{document}